%% file: main.tex
\documentclass{article}

\usepackage[preprint]{nips_2018}

\usepackage[utf8]{inputenc} 
\usepackage[T1]{fontenc}    
\usepackage{hyperref}       
\usepackage{url}            

\usepackage{booktabs}       
\usepackage{amsmath}        
\usepackage{amsfonts}       
\usepackage{nicefrac}       
\usepackage{microtype}      
\usepackage{siunitx}        
\usepackage{multicol}

\usepackage{graphicx}
\usepackage{subcaption}
\newsavebox{\imagebox}

\setcitestyle{authoryear,brackets=round}

\title{Inverse reinforcement learning for video games}

\author{
  Aaron Tucker \\
  UC Berkeley \\
  \texttt{aarondtucker@gmail.com}
  \And
  Adam Gleave \\ 
  UC Berkeley \\
  \texttt{gleave@berkeley.edu}
  \And
  Stuart Russell \\
  UC Berkeley \\
  \texttt{russell@cs.berkeley.edu}
}

\begin{document}

\maketitle

\begin{abstract}
Deep reinforcement learning achieves superhuman performance in a range of video game environments, but requires that a designer manually specify a reward function.
It is often easier to provide demonstrations of a target behavior than to design a reward function describing that behavior.
Inverse reinforcement learning (IRL) algorithms can infer a reward from demonstrations in low-dimensional continuous control environments, but there has been little work on applying IRL to high-dimensional video games.
In our CNN-AIRL baseline, we modify the state-of-the-art adversarial IRL (AIRL) algorithm to use CNNs for the generator and discriminator.
To stabilize training, we normalize the reward and increase the size of the discriminator training dataset.
We additionally learn a low-dimensional state representation using a novel autoencoder architecture tuned for video game environments.
This embedding is used as input to the reward network, improving the sample efficiency of expert demonstrations.
Our method achieves high-level performance on the simple Catcher video game, substantially outperforming the CNN-AIRL baseline.
We also score points on the Enduro Atari racing game, but do not match expert performance, highlighting the need for further work.
\end{abstract}

\input{intro}

\input{relatedwork}

\input{airl}

\input{varred}

\input{experiments}

\input{conclusions}

\subsubsection*{Acknowledgments}
\makeatletter
\if@submission
Removed for double-blind submission.
\else
This work was supported by the Center for Human-Compatible AI and the Open Philanthropy Project, the Future of Life Institute and the Leverhulme Trust. We would like to thank Sam Toyer, Lawrence Chan, Matthew Rahtz and Daniel Filan for comments on an earlier draft.
\fi
\makeatother

\bibliography{main}{}

\end{document}

%% file: intro.tex
\section{Introduction}
Deep reinforcement learning has achieved great success at optimizing known reward functions in a range of challenging environments.
However, designing an appropriate reward function is difficult even for experienced engineers, and is impossible for end-users.
A natural solution is to learn the reward function from human demonstrations: inverse reinforcement learning (IRL)~\citep{irl}.
We seek to scale IRL algorithms to MDPs with high-dimensional state spaces and discontinuous dynamics, such as Atari and PyGame environments~\citep{ale,ple}.

Recent deep IRL algorithms have achieved good performance on a variety of continuous control tasks~\citep{gcl,airl}.
However, no IRL algorithm has been able to attain human-level performance on Atari games.
This is remarkable considering contemporary deep RL algorithms outperform humans on 70\% of Atari games~\citep{rainbow}.
This suggests we currently have substantially more powerful tools for optimizing known objectives than for learning the objectives themselves.
Yet reinforcement learning is only as strong as its weakest link: both a powerful optimizer and an accurate reward function are needed for good results.
Accordingly, closing this capability gap between reward optimization and reward learning is critical to support human-centric AI applications.

A key challenge is that video game environments have substantially higher-dimensional state spaces compared to continuous control tasks. 
For example, Atari games have a $\num{28224}$-dimensional state space.\footnote{We assume the standard preprocessing to reduce dimensionality, giving a state space of $84 \times 84 \times 4 = 28,224$~\citep{dqn}.}
By contrast, even the challenging Humanoid Mujoco environment has only $376$ dimensions~\citep{humanoidmujoco}, with most continuous control tasks having far smaller state spaces.

Additionally, the reward function in video games is often discontinuous.
For example, a shooter game gives positive reward when a shot touches a target, but no reward if the shot is a pixel away from the target.
By contrast, continuous control tasks typically have smooth reward functions, such as the distance from a goal or forward velocity.

Related to this, expert demonstrations tend to be multimodal.
For example, an optimal policy might dodge a target by veering sharply to the left or the right.
The direction chosen is unimportant, and could be random or depend on small details (such as whether the agent is already slightly to the left or right of the target).
A naive reward learner, seeing demonstrations both to the left and the right in similar states, might assign similar reward to both the left and right action.
But the resulting policy would not travel far in either direction, and might consequently crash into the target.
It is therefore necessary for a sharp decision boundary to be learned, which is typically harder to represent.


Our approach is based on the adversarial training method pioneered by GAN Guided Cost Learning (GAN-GCL) and adversarial IRL~\citep{gan-gcl,airl}.
As a baseline, we extended adversarial IRL to support discrete action spaces and replaced the policy and reward networks with CNNs \citep{cnn}.
We found this baseline never achieves better than random performance in Atari games, although it is able to score some points in a simple PyGame environment, Catcher.
This is unsurprising as adversarial training is often unstable, especially in high-dimensional environments.
Indeed, a key benefit of adversarial IRL over GAN-GCL is the variance reduction from discriminating state-action pairs rather than entire trajectories.

We use two methods to scale IRL to video games.
First, we train the discriminator on samples from previous versions of the generator (policy) in addition to the current version.
This is similar to a variance-reduction technique used in guided cost learning \citep{gcl}, but omitted from adversarial IRL.
Second, we train an autoencoder on environment frames collected via random exploration.
Conventional autoencoder designs tend to neglect small, difficult to model objects such as a moving ball.
We propose a novel autoencoder architecture based on a mixture of Gaussians model that is better suited to video game environments.
The resulting low-dimensional embedding is used as an input to the reward network (discriminator).
The policy network (generator) continues to receive raw image inputs.

Together, we find these techniques are able to stabilize training to achieve near expert-level performance on Catcher and to score points in Enduro, an Atari racing game.
These results support the hypothesis that GAN-based inverse reinforcement learning methods are capable of learning rewards in video games provided the variance can be managed.
In further work, we plan to investigate other methods for stabilizing GAN training in order to extend our approach to a broader class of games.

%% file: relatedwork.tex
\section{Related work}

Inverse reinforcement learning (IRL) was first described by \citet{irl}. Early work assumed optimal demonstrations and a reward function that is linear in a known set of features. The key challenge was reward ambiguity: given an optimal policy for some MDP, there are many reward functions that could have led to this policy. \citet{irl2004} developed a method that, in the limit of infinite data, recovers a policy that obtains the same reward as the expert on the original MDP. However, it will not surpass the expert in performance, and the value obtained may be arbitrarily bad on MDPs with the same reward but different transition dynamics.

The next key wave of work came in the form of Bayesian IRL and Maximum Entropy IRL. Bayesian IRL embraced reward ambiguity, inferring a posterior distribution over rewards rather than committing to a given reward function~\citep{bayesianirl}. By contrast, Maximum Entropy IRL returns a reward function that matches the expected feature counts, favoring rewards that lead to a higher-entropy stochastic policy~\citep{maxent,maxcausalent}. The feature matching constraint gives the same guarantee as \citet{irl2004}, while maximizing entropy improves generalization to environments with different dynamics.

In addition, both Bayesian and Maximum Entropy IRL relaxed the optimal demonstrations assumption, instead modeling the expert as being Boltzmann rational. That is, the probability of the expert taking an action is proportional to the exponential of the $Q$-value:
\begin{equation*}
\pi^E(a \mid s) \propto \exp\left(Q^*(s,a)\right).
\end{equation*}
A policy optimizing the resulting reward function can therefore do \textit{better} than the demonstrations originally provided, a key advantage over previous work.

Most recent work has built on Maximum Entropy IRL since it is amenable to computationally efficient implementations.
The original Maximum Entropy IRL algorithm assumed known transition dynamics, a finite state space and a linear reward over features.
Successive developments have relaxed these restrictions.
Relative Entropy IRL scales to MDPs with infinite state spaces and unknown transition dynamics, but continues to assume a linear reward~\citep{relentirl}.
By contrast, deep IRL can learn a non-linear reward function, but requires a finite state space with known dynamics~\citep{deepirl}.
Guided Cost Learning (GCL) pioneered a hybrid of these two approaches, and was the first algorithm able to learn non-linear reward functions over an infinite state space with unknown transition dynamics~\citep{gcl}.
It was quickly noticed that GCL was related to GAN training~\citep{gan-gcl}, which was then exploited with adversarial IRL~\citep{airl}.

Despite this large body of work, to the best of our knowledge the only attempt at applying IRL to video games is by~\citet{uchibe}. This approach classifies state transitions as expert or non-expert using logistic regression. The classifier is then used as a reward function to train a deep RL algorithm. Their experiments show this algorithm rarely outperforms the behavioral cloning baseline, underscoring the need for further work in this area. Our method differs by training the policy and reward jointly in an adversarial fashion, and by employing a variety of variance reduction techniques.

Imitation learning from visual data has a long history in mobile robotics, including autonomous driving~\citep{alvinn,nvidiacar} and quadcopter control~\citep{daggermonocular}. Recent work has made progress in sample efficiency~\citep{oneshotvisim} and learning from third-person observations~\citep{imfromobs,thirdperson}. However, there is little work involving imitation learning on video games.

To the best of our knowledge, deep $Q$-learning from demonstrations (DQfD) was the first work to use demonstrations in Atari games~\citep{dqfd}. However, DQfD uses demonstrations to bootstrap reinforcement learning against a known reward function, rather than learning the reward function from demonstrations. The authors also evaluated a simple behavioral cloning baseline, which predictably performed poorly on the majority of games. \citet{grandchallenge} collected a human dataset on Atari games and independently evaluated behavioral cloning, obtaining similar negative results. Recent work by \citet{hardexploration} imitates expert game play from YouTube videos, outperforming average human performance on three `hard exploration' Atari games.

\citet{drlfromhumans} applied active preference learning to Atari games, asking users to select the best of two trajectories generated from an ensemble of policies. The policies were trained to maximize a reward function that was being learned from user feedback in an iterative process. \citeauthor{drlfromhumans} achieve good performance, matching a direct RL approach in most cases and in one environment even outperforming it. Our work differs from this approach by learning directly from demonstrations, avoiding the need for interactive user feedback.

%% file: airl.tex
\section{CNN-AIRL}
\label{sec:cnn-airl}
We build on adversarial IRL (AIRL), an algorithm achieving state-of-the-art performance on simulated robotics tasks~\citep{airl}. The reference implementation of AIRL assumes a continuous action space and uses a fully-connected policy and reward network, a poor fit for high-dimensional image inputs~\citep{airlcode}. We developed CNN-AIRL as a baseline, making the minimal set of modifications needed to run AIRL on video games. In this section we summarize CNN-AIRL, deferring discussion of more substantial modifications to section~\ref{sec:var-red}.

Adversarial IRL formulates the inverse reinforcement learning problem as a GAN~\citep{gan}. We learn a reward function $f_{\theta}(s,a)$ for taking action $a$ in state $s$ and a stochastic policy $\pi(a \mid s)$. The policy is the generator, and is trained using forward RL on the reward function $f_{\theta}(s,a)$. The discriminator is restricted to have the special form:
\begin{equation*}
D_{\theta}(s,a) = \frac{\exp\left(f_{\theta}(s,a)\right)}{\exp\left(f_{\theta}(s,a)\right) + \pi(a \mid s)}.
\end{equation*}
The discriminator is trained via logistic regression to distinguish between expert demonstrations and background samples from the generator. At optimality, $f^*(s,a) = \log \pi^*(a \mid s) = A^*(s,a)$, the advantage function of the optimal policy~\citep[appendix A]{airl}.


AIRL has previously only been applied to simulated robotics tasks, with low-dimensional inputs and continuous action spaces. In order to scale to image inputs, we substituted a convolutional neural network for the multilayer perceptron in the reward and policy model. For the policy network, we use the architecture described in \citet{dqn}. We base the reward network on the same architecture, but modify it to use a batch normalized leaky ReLU \citep{leakyrelu} as used in DCGAN~\citep{dcgan}.

The reward network therefore consists of three convolutional layers, followed by two fully-connected layers. For state-only reward networks $R(s)$, we make no further modifications. For a reward network $R(s,a)$ taking states and actions as input, we concatenate the one-hot coded action vector to the CNN output, before the fully-connected layers.

Our final modification is to use Proximal Policy Optimization (PPO)~\citep{ppo} to train the policy network instead of Trust-Region Policy Optimization (TRPO)~\citep{trpo}, used in the original AIRL implementation. We selected PPO since it outperforms TRPO on Atari games (in tests where the reward is known), and is simpler to implement.

%% file: varred.tex
\section{Stabilizing training}
\label{sec:var-red}

In the previous section, we described CNN-AIRL: a lightly modified version of AIRL that is able to operate in video game environments. However, unsurprisingly the performance of this algorithm proved underwhelming in tests. In this section, we propose several extensions to this baseline approach.

\subsection{Dataset expansion}
Adversarial IRL (AIRL) trains the discriminator to distinguish between expert demonstrations, and background samples from the generator.
At each discriminator training step, a dataset of state-action pairs is sampled from both the demonstrations and background samples.
Gradient descent is then performed for several iterations to minimize the cross-entropy loss.
If the dataset is too small, the discriminator may overfit.
The reward provided to the generator will then have low information content, and neither the discriminator nor the generator will converge.

In the original implementation of AIRL, the most recent rollout of the generator is used for background samples, with an equal number of samples taken from the demonstrations.
However, the length of generator rollouts is typically fairly small: we use a rollout of $1024$ timesteps.
This dataset size is too small for discriminator training.
In general, there is no reason why the appropriate dataset size for inverse RL should be the same as the number of timesteps of forward RL experience.

In dataset expansion, we increase the number of background samples by using the last $k$ rollouts of forward RL, similar to an approach employed in guided cost learning~\citep{gcl}. In our experiments, we implement dataset expansion by running $k$ steps of forward RL for each discriminator training step. However, it is possible to vary the ratio of generator and discriminator training separately; we leave separate tuning of these hyperparameters to further work.

\subsection{Reward normalization}
In adversarial IRL, we alternate between generator and discriminator training which update the policy and reward networks respectively. During generator training, we run forward RL to update the policy network based on rewards predicted by the reward network. Forward RL algorithms typically assume a stationary reward: true in most environments, but not when the reward network is being updated.

In particular, policy gradient approaches such as TRPO or PPO typically take gradient steps depending on an estimate of the advantage function. The advantage function in turn depends on an estimate of the value of each state. The predicted value is based on rewards output by previous reward networks, that may be substantially different to the current reward.

Adversarial training necessarily involves a non-stationary component, so there is no simple solution to this problem. However, since the optimal policy is invariant under positive affine transformations of the reward function, we can at least fix the mean and standard deviation of reward over time. Specifically, we center and rescale rewards using the mean and standard deviation of rewards on the last sample of trajectories used to train the discriminator. Note this still allows a policy to achieve above-mean reward during forward RL training, as the reward is only re-normalized after discriminator training.

\subsection{Pixel-class autoencoders}
Video games are challenging in large part due to their high dimensional state space. Deep RL algorithms are able to learn good policies, but require millions of frames of experience. However, collecting this many human demonstrations would be slow and prohibitively expensive. Our IRL algorithm therefore needs to be orders of magnitude more sample efficient than RL algorithms when it comes to human demonstrations.

Although the raw input from a video game is high dimensional, the intrinsic dimensionality of the task is much lower. For example, in Pong it is sufficient to represent the coordinates and velocity of the paddles and ball, and the current score of each player. Crucially, it should be possible to learn a good representation through only unsupervised exploration of the environment.

We use an autoencoder to learn a low-dimensional embedding for frames collected by random exploration in a video game. This embedding forms the input to the reward network, allowing it to learn from a small number of expert demonstrations. The policy network continues to receive direct visual input, since we are not concerned with the sample efficiency of unsupervised rollouts.

We found many standard autoencoder methods failed to capture critical details such as the position of a ball or paddle, with the object often vanishing entirely in the decoded image. This motivated us to develop a new architecture which is better suited to reconstructing frames from small embedding vectors. Our key insight is that Atari and other simple video games use a small number of discrete colors in the display. 

We model each pixel as being drawn from a mixture of Gaussians, with each Gaussian corresponding to a particular type of object. Our pixel-class CNN outputs class logits $z_{ijk}$ for each pixel $(i,j)$, rather than making a direct prediction of the pixel value. The class label $c_{ij}$ is sampled from $\mathrm{softmax}(\mathbf{z}_{ij})$. The grayscale value $x_{ij}$ for pixel $(i,j)$ is drawn from the Gaussian $\mathcal{N}(\mu_{c_{ij}}, \sigma_{c_{ij}}^2)$. The $\mu_k$ and $\sigma_k$ are variables that are jointly learned with the logits $z_{ijk}$. We train the CNN to maximize the likelihood of the data under this observation model.

Prior work in image segmentation has used a mixture of Gaussians observation model~\citep{uai97}. However, this previous work assumes a constant class probability for each pixel and uses an EM algorithm to infer the model parameters. By contrast, we use a CNN to output the class logits based on the image input, and perform maximum likelihood estimation via gradient descent.

%% file: experiments.tex
\section{Experiments}
\begin{figure}[p]
    \savebox{\imagebox}{\includegraphics[width=0.48\textwidth]{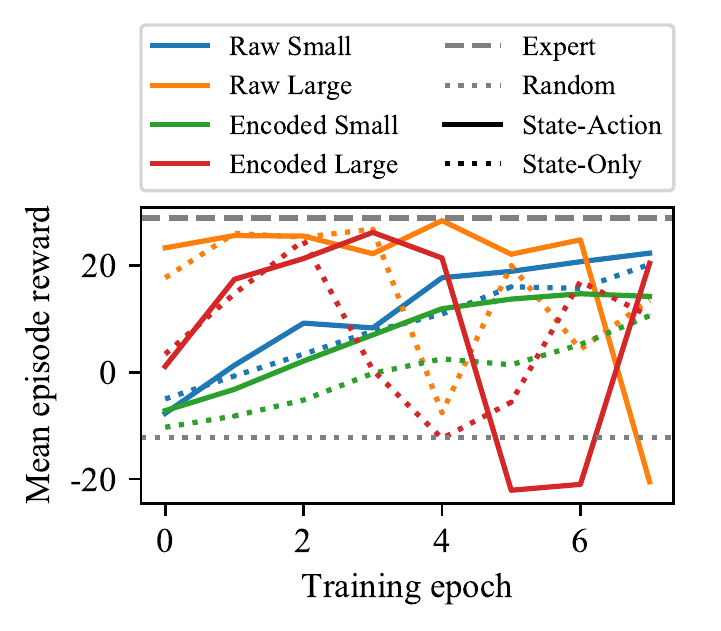}}%
    \begin{subfigure}[t]{0.25\textwidth}
        \raisebox{\dimexpr.5\ht\imagebox-.5\height}{%
            \includegraphics[width=\textwidth]{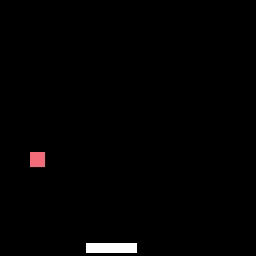}}%
        \caption{Screenshot from random exploration.}
        \label{fig:catcher:frame}
    \end{subfigure}
    \hspace{2em}
    \begin{subfigure}[t]{0.7\textwidth}
        \centering
        \usebox{\imagebox}    
        \caption{Mean episode reward of IRL policy. Colors denote whether the discriminator input was raw images or an encoding, and whether the batch size was small or large. Solid or dashed line indicates if the discriminator also received an action input or just the state.}
    \label{fig:catcher:irl}
    \end{subfigure}
    \label{fig:catcher}
    \caption{IRL on Catcher.}
\end{figure}

\begin{figure}[p]
    \savebox{\imagebox}{\includegraphics[width=0.48\textwidth]{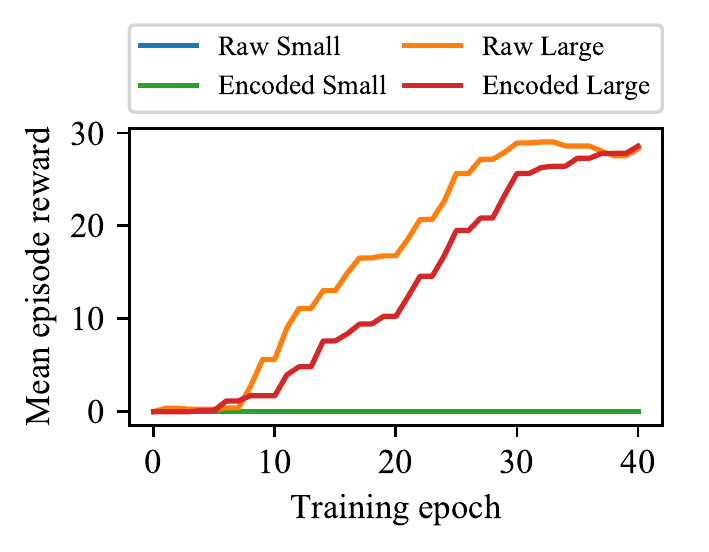}}%
    \begin{subfigure}[t]{0.25\textwidth}
        \raisebox{\dimexpr.5\ht\imagebox-.5\height}{%
            \includegraphics[width=\textwidth]{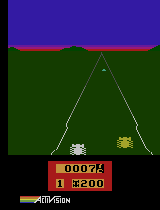}}%
        \caption{Screenshot from random exploration.}
        \label{fig:enduro:frame}
    \end{subfigure}
    \hspace{2em}
    \begin{subfigure}[t]{0.7\textwidth}
        \centering
        \usebox{\imagebox}    
        \caption{Mean episode reward of IRL policy. Colors denote whether the discriminator input was raw images or an encoding, and whether the batch size was small or large. In this test, the discriminator always receives an action input.}
    \label{fig:enduro:irl}
    \end{subfigure}
    \label{fig:enduro}
    \caption{IRL on Enduro.}
\end{figure}

\begin{figure}[p]
    \centering
    \begin{subfigure}[t]{0.45\textwidth}
        \includegraphics[width=\textwidth]{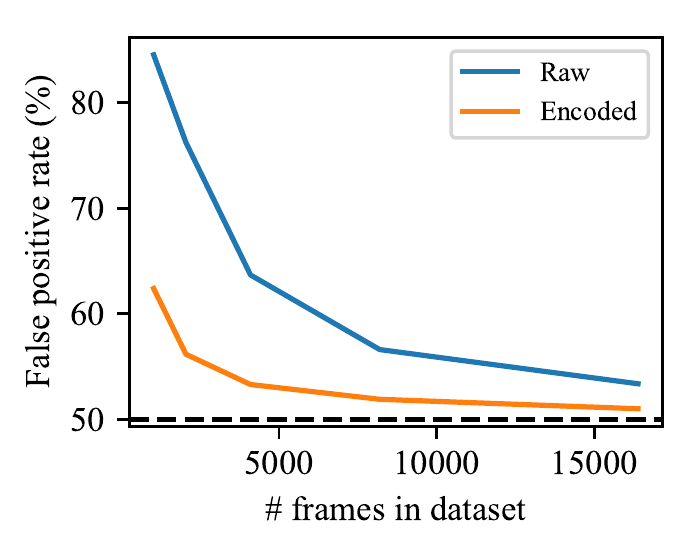}
        \caption{Catcher.}
        \label{fig:catcher:overfitting}
    \end{subfigure}
    \begin{subfigure}[t]{0.45\textwidth}
        \includegraphics[width=\textwidth]{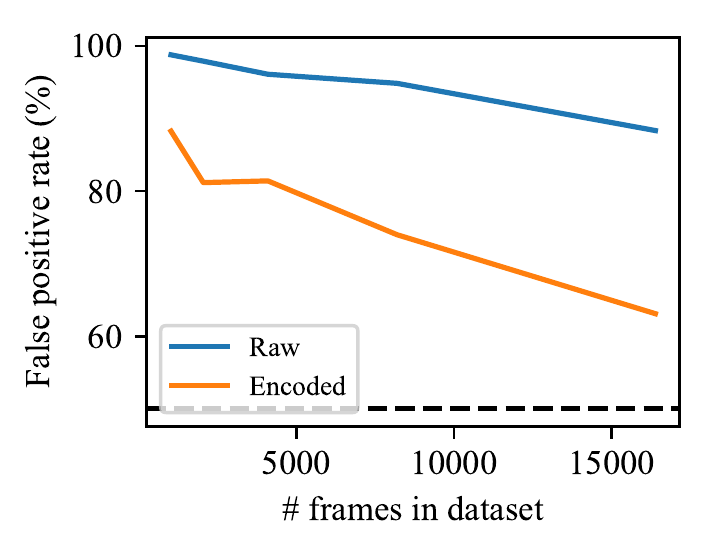}
        \caption{Enduro.}
        \label{fig:enduro:overfitting}
    \end{subfigure}
    \caption{False positive rate for the discriminator on unseen samples from the expert policy. A rate in excess of 50\% is indicative of overfitting.}
    \label{fig:overfitting}
\end{figure}

We evaluated our method on Catcher, a simple PyGame environment, and Enduro, an Atari racing game. We generated synthetic expert demonstrations from a policy trained with PPO on the ground-truth reward function. We train our IRL algorithm on eight trajectories sampled from this expert policy, learning a joint reward-policy pair. We evaluate on an apprenticeship learning metric: the ground-truth reward obtained by the resulting policy. To elucidate the contribution of the different modifications we made to adversarial IRL, we also evaluate how the variance of the discriminator varies with dataset size, and compare our pixel-class autoencoder to standard techniques.

\subsection{Performance on Catcher}
Catcher is a PyGame environment involving a paddle and falling blocks (example frame in fig~\ref{fig:catcher:frame}.)
A reward of $+1$ is received when a block touches the paddle, and $-1$ if it touches the ground.
We report the results of an ablation study in fig~\ref{fig:catcher:irl}.
Our best methods achieve performance comparable to the expert.
The greatest improvement comes from dataset expansion, with the \textit{Raw Large} variant outperforming the baseline CNN-AIRL \textit{Raw Small} version.
However, there is no benefit from using an autoencoder in this environment, with the \textit{Encoded} variants achieved comparable reward to the \textit{Raw} versions, perhaps because the visual input is so simple.
There is also little difference between a state-action and state-only input to the reward network.
All variants exhibit high variance.

\subsection{Performance on Enduro}
Catcher provides a useful proof-of-concept environment, but is drastically simpler than most video games. As a challenge environment, we evaluate on Enduro, an Atari racing game with more complex dynamics (acceleration, steering, collisions, road curvature) and visual distractions (both the background and cars change color over time). We report our results in figure~\ref{fig:enduro:irl}. As expected, the baseline \textit{Raw Small} variant fails to make progress, scoring zero points (as does the random policy). Both our \textit{Raw Large} and \textit{Encoded Large} variants score some points. Qualitatively, the policies appear to have learnt that acceleration is rewarding and are able to steer to stay on the road, but have not learnt to avoid colliding with other vehicles. Although the performance is substantially below the expert policy, which scores over 400 points, these results suggest the framework is capable of learning appropriate reward functions. Further improvements in stabilizing the adversarial training will be needed to yield full performance.

\subsection{Discriminator overfitting}
We have seen in the previous sections that use of a larger dataset when training the discriminator improve IRL performance. These modifications were originally motivated by a hypothesis that the discriminator was prone to overfit to the expert demonstrations. We test this by presenting the discriminator with unseen trajectories sampled from the same expert policy used to generate the in-sample expert demonstrations. Ideally, the discriminator false positive rate would be 50\%: any value in excess of this indicates it has overfit to the particular samples.

In figure~\ref{fig:overfitting}, we report results on Catcher and Enduro, both for raw image input and a low-dimensional encoding. Use of a larger dataset size drastically decreases the false positive rate in both environments. On Catcher, using a low-dimensional encoding decreases the false positive rate with small datasets, but has little effect with larger datasets. 

On Enduro, the low-dimensional encoding substantially decreases the false positive rate.
Remarkably, the false positive rate for the encoding with $\num{1024}$ frames (the smallest dataset tested) is lower than for the raw input with $\num{16834}$ frames (the largest dataset). Despite this, we saw in the previous section that dataset size has a greater effect on policy performance than the use of an encoding, suggesting that a larger dataset may have other beneficial effects not captured in this test.

\subsection{Pixel-class autoencoder performance}
\newcommand\compare[1]{
 \includegraphics[width=.25\textwidth]{figures/pong/encoder/input/#1.png} &
 \includegraphics[width=.25\textwidth]{figures/pong/encoder/normal/#1.png} &
 \includegraphics[width=.25\textwidth]{figures/pong/encoder/pixel_class/#1.png}
}
\begin{figure}
    \centering
    \begin{subfigure}{0.49\textwidth}
    \begin{tabular}{ccc}
    Input &  Normal &  Pixel-Class\\ \hline
    \compare{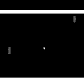}\\
    \compare{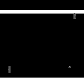}\\
    \compare{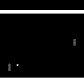}\\
    \compare{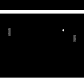}\\
    \end{tabular}
    \end{subfigure}
    \begin{subfigure}{0.49\textwidth}
    \begin{tabular}{ccc}
    Input &  Normal &  Pixel-Class\\ \hline
    \compare{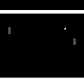}\\
    \compare{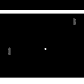}\\
    \compare{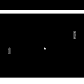}\\
    \compare{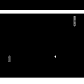}\\
    \end{tabular}
    \end{subfigure}
    \caption{Qualitative comparison of decoded Pong frames from a conventional and our pixel-class autoencoder.}
    \label{fig:autoencoder-pong-imgs}
\end{figure}

The previous section showed that using an autoencoder reduces discriminator overfitting. In figure~\ref{fig:autoencoder-pong-imgs}, we directly compare decoded images from our pixel-class autoencoder to images produced by a conventional autoencoder on Pong. While both methods obtain reasonably high-fidelity, our pixel-class autoencoder achieves a substantially sharper rendering of the moving ball, with the ball in one case almost disappearing in the conventional design (row 3, left). We also find the pixel-class autoencoder to have comparable mean-squared error to the conventional autoencoder, despite this not being a design objective of our autoencoder.

Our primary focus is on inverse reinforcement learning, not autoencoder design, and we believe better autoencoder designs exist; we postpone discussion of this to the further work section. Of course, any autoencoder improvement would tend to improve the performance of our IRL algorithm, the primary contribution of this paper. 

%% file: conclusions.tex
\section{Discussion}

\subsection{Summary}
We have developed the first adversarial inverse reinforcement learning algorithm applicable to video game environments. Two key modifications were needed to achieve stable training: normalizing the reward to avoid drift over time, and collecting background samples from multiple iterations of forward RL. We additionally learn a low-dimensional embedding with an autoencoder to improve the sample efficiency on expert demonstrations. With these modifications, we are able to achieve near expert-level performance on Catcher, a simple PyGame environment. Our code and experimental results are available at 
\makeatletter
\if@submission
---removed for double-blind submission---.
\else
\url{https://github.com/HumanCompatibleAI/atari-irl}.
\fi
\makeatother

\subsection{Limitations and future work}
Our performance on Atari games improves compared to the baseline vanilla CNN-AIRL approach, but remains substantially below expert-level on the games we evaluated on. This highlights the need for further work to scale inverse reinforcement learning to Atari games.

Increasing the number of background samples reduces, but does not eliminate, discriminator overfitting. In concurrent work, \citet{variationalbottleneck} developed a variational discriminator bottleneck that stabilizes a range of adversarial learning tasks, including adversarial IRL on simple point-mass environments. Application of this method might further reduce overfitting, improving performance in the Atari domain.

Use of an autoencoder to learn a low-dimensional representation of the state substantially improved performance in our experiments. However, autoencoders are optimized to recover the original image, which includes many details that are irrelevant to the reward. We expect the recent Causal InfoGAN method that takes into account the sequential nature of the data would learn a more suitable embedding~\citep{causalinfogan}. 

There is also scope for improving the data used to train the autoencoder. We currently collect frames via random exploration, but this is unlikely to reach many important states. It would be better to use a more systematic approach for unsupervised exploration, such as Diversity is All You Need~\citep{diayn}. Alternately, one could periodically re-train the autoencoder using frames collected from rollouts of the policy.

Currently, we use the Proximal Policy Optimization (PPO) RL algorithm for generator training~\citep{ppo}. However, $Q$-learning methods such as DQN typically attain higher reward on Atari games than policy-gradient algorithms such as PPO. Unfortunately, $Q$-learning methods are likely to be particularly sensitive to the reward function changing after each discriminator training step. A resolution to this, perhaps extending our reward normalization heuristic, could substantially improve performance.